\documentclass[sigconf, acmart]{acmart}

\AtBeginDocument{%
  \providecommand\BibTeX{{%
    \normalfont B\kern-0.5em{\scshape i\kern-0.25em b}\kern-0.8em\TeX}}}

\copyrightyear{2022}
\acmYear{2022}
\setcopyright{acmlicensed}\acmConference[KDD '22]{Proceedings of the 28th ACM SIGKDD Conference on Knowledge Discovery and Data Mining}{August 14--18, 2022}{Washington, DC, USA}
\acmBooktitle{Proceedings of the 28th ACM SIGKDD Conference on Knowledge Discovery and Data Mining (KDD '22), August 14--18, 2022, Washington, DC, USA}
\acmPrice{15.00}
\acmDOI{10.1145/3534678.3539472}
\acmISBN{978-1-4503-9385-0/22/08}

\usepackage{multirow}
\usepackage{booktabs}
\usepackage{natbib}
\usepackage{titlesec}
\usepackage[ruled,vlined]{algorithm2e}
\usepackage[utf8]{inputenc}
\usepackage{amssymb,amsmath,caption}
\usepackage{lipsum}
\usepackage{listings}
\usepackage{extarrows}
\usepackage{subfigure}
\usepackage{xspace}
\usepackage{graphicx}
\usepackage{makecell}
\usepackage{ifthen}
\usepackage{xifthen}
\usepackage{wrapfig}
\usepackage{amsthm}
\usepackage{stmaryrd}
\usepackage{threeparttable}
\usepackage{enumerate}

\usepackage[dvipsnames]{xcolor}
\usepackage{tikz}
\usetikzlibrary{backgrounds}
\usetikzlibrary{arrows,shapes}
\usetikzlibrary{tikzmark}
\usetikzlibrary{calc}

\usepackage{mathtools, nccmath}
\usepackage{comment}

\usepackage{blindtext}

\usepackage{tcolorbox}
\usepackage{enumerate}
\usepackage{enumitem}

\usepackage{tikz}
\usetikzlibrary{arrows,shapes,positioning,shadows,trees,mindmap}
\usepackage[edges]{forest}
\usetikzlibrary{arrows.meta}
\colorlet{linecol}{black!75}
\usepackage{xkcdcolors} 

\usepackage{tikz}
\usetikzlibrary{backgrounds}
\usetikzlibrary{arrows,shapes}
\usetikzlibrary{tikzmark}
\usetikzlibrary{calc}

\colorlet{mhpurple}{Plum!80}

\newcommand\condent{\raisebox{-2pt}{\includegraphics[width=0.8em]{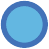}}}
\newcommand\interent{\raisebox{-2pt}{\includegraphics[width=0.8em]{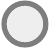}}}
\newcommand\maskent{\raisebox{-2pt}{\includegraphics[width=0.8em]{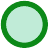}}}

\newcommand{\method}[0]{\textrm{kgTransformer}\xspace}
\newcommand{\vpara}[1]{\vspace{0.07in}\noindent\textbf{#1}\xspace}

\newcommand{\beqn}[1]{{\small\begin{eqnarray}#1\end{eqnarray}}}

\newcommand{\shiyu}[1]{\textbf{\color{orange}[(shiyu: #1 )]}}

\newcommand{\tohide}[1]{} 

\begin{document}

\title{Mask and Reason: Pre-Training Knowledge Graph Transformers for Complex Logical Queries}

\author{Xiao Liu}
\authornote{The authors contributed equally to this research.}
\affiliation{Tsinghua University}
\email{liuxiao21@mails.tsinghua.edu.cn}

\author{Shiyu Zhao}
\authornotemark[1]
\affiliation{Tsinghua University}
\email{sy-zhao19@mails.tsinghua.edu.cn}

\author{Kai Su}
\authornotemark[1]
\affiliation{Tsinghua University}
\email{suk19@mails.tsinghua.edu.cn}

\author{Yukuo Cen}
\affiliation{Tsinghua University}
\email{cyk20@mails.tsinghua.edu.cn}

\author{Jiezhong Qiu}
\affiliation{Tsinghua University}
\email{jiezhongqiu@outlook.com}

\author{Mengdi Zhang}
\affiliation{Meituan-Dianping Group}
\email{zhangmengdi02@meituan.com}

\author{Wei Wu}
\affiliation{Meituan-Dianping Group}
\email{wuwei30@meituan.com}

\author{Yuxiao Dong}
\authornotemark[2]
\affiliation{Tsinghua University}
\email{yuxiaod@tsinghua.edu.cn}

\author{Jie Tang}
\authornote{Jie Tang and Yuxiao Dong are the corresponding authors.}
\affiliation{Tsinghua University}
\email{jietang@tsinghua.edu.cn}




\renewcommand{\shortauthors}{Xiao Liu et al.}
\begin{abstract} \label{sec:abs}

Knowledge graph (KG) embeddings have been a mainstream approach for reasoning over incomplete KGs. 
However, limited by their inherently shallow and static architectures, they can hardly deal with the rising focus on complex logical queries, which comprise logical operators, imputed edges, multiple source entities, and unknown intermediate entities. 
In this work, we present the Knowledge Graph Transformer (\method)\footnote{The code is available at \url{https://github.com/THUDM/kgTransformer}} with masked pre-training and fine-tuning strategies. 
We design a KG triple transformation method to enable  Transformer  to handle KGs, which is further strengthened by the Mixture-of-Experts (MoE) sparse activation. 
We then formulate the complex logical queries as masked prediction and introduce a two-stage masked pre-training strategy to improve transferability and generalizability.
Extensive experiments on two benchmarks demonstrate that \method can consistently outperform both KG embedding-based baselines and advanced encoders on nine in-domain and  out-of-domain reasoning tasks. 
Additionally, \method can reason with explainability via providing the full reasoning paths to interpret given answers.

\end{abstract}



\tohide{

}



\maketitle

\section{INTRODUCTION}
\label{sec:intro}

Knowledge graphs (KGs) store and organize human knowledge about the factual world, such as the human-curated  Freebase~\cite{bollacker2008freebase} and Wikidata~\cite{vrandevcic2014wikidata} as well as the semi-automatic constructed ones---NELL~\cite{carlson2010toward} and Knowledge Vault~\cite{dong2014knowledge}. 
Over the course of KGs' development, representation learning for querying KGs is one of the fundamental problems. 
Its main challenge lies in the incomplete knowledge and inefficient queries. Web-scale KGs are known to suffer from missing links~\cite{west2014knowledge}, and the specialized querying tools such as SPARQL cannot deal well with it. 

Knowledge graph embeddings (KGEs), which aim at embedding entities and relations into low-dimensional continuous vectors, have thrived in the last decade~\cite{bordes2013translating,trouillon2016complex}. 
Specifically, KGEs have found wide adoptions in the simple KG completion problem ($h, r,?$), which features a single head entity $h$, a relation $r$ and the missing tail entity. 
However, real-world queries can be more complicated with imputed edges, multiple source entities, Existential Positive First-Order (EPFO) logic, and unknown intermediates, namely, the \textit{complex logical queries}.

\begin{figure} 
    \centering
    \includegraphics[width=1.0\linewidth]{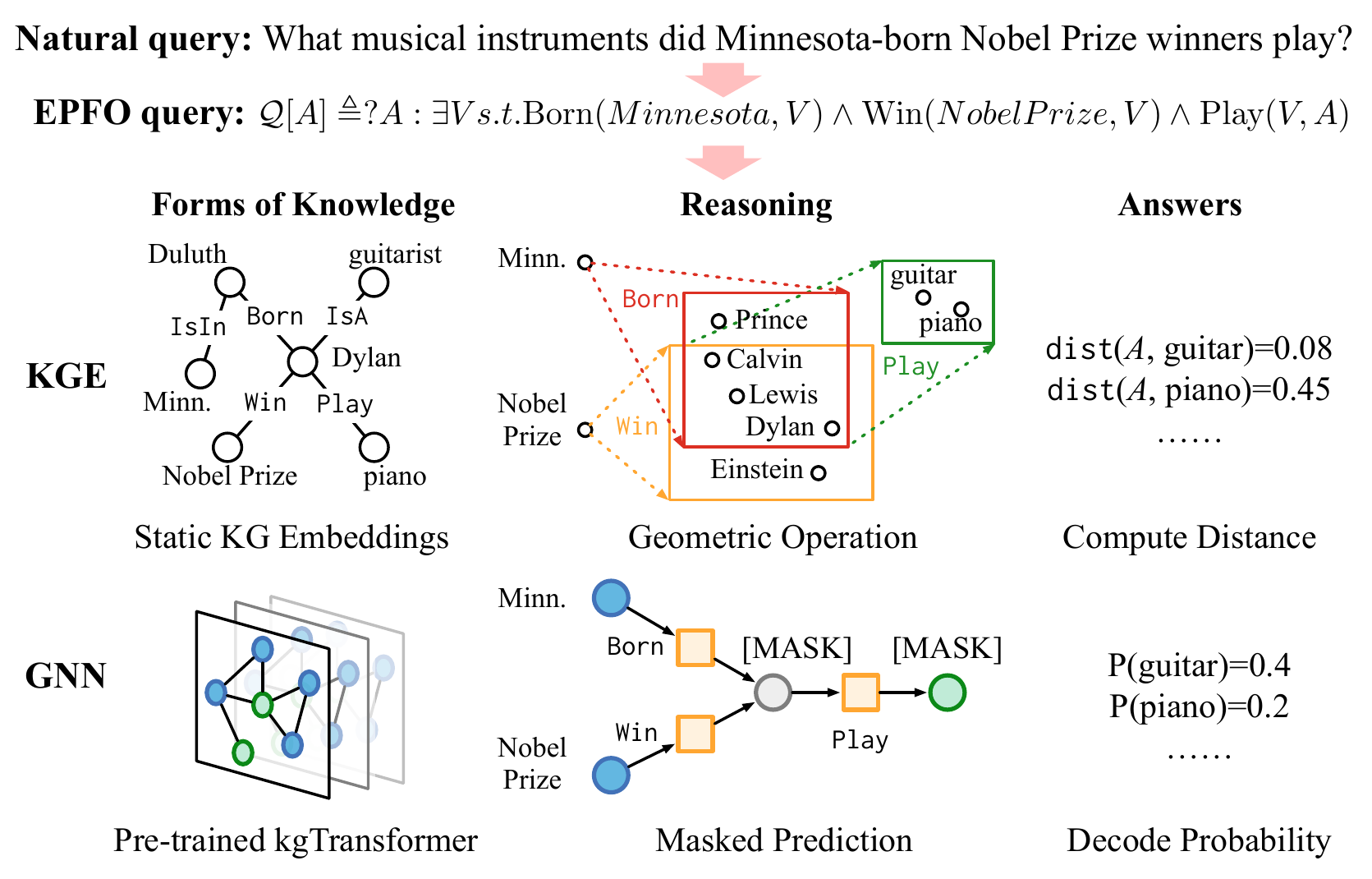}
    \caption{EPFO query reasoning: KGE-based reasoners vs. Pre-trained \method. \textmd{The masked prediction training can endow GNNs with natural capability to answer EPFO queries.}}
    \label{fig:intro}
\end{figure}

Figure~\ref{fig:intro} illustrates a decomposed query graph for the complex logical query ``What musical instrument did Minnesota-born Nobel Prize winner play?''. 
The query composes two source entities (``Minnesota'' \& ``Nobel Prize'', denoted as ``\condent{}''), First-Order logic operator conjunction ($\wedge$), unknown intermediate entities (``people that were born in Minnesota and won Nobel Prize'', denoted as ``\interent{}''), unknown target entities (``musical instruments they may play'', denoted as ``\maskent{}''), and potential missing edges.
A major challenge of answering such  query is the exponential complexity along its growing combinations of hops and logical operators. 
Additionally, the rich context information surrounding multiple entities and relations in a single query should also be taken into account during reasoning.

Consequently, such query goes beyond the capability of existing KGE-based approaches. 
First, most KGEs' architectures are shallow and use static vectors, limiting their expressiveness and capacity to capture massive patterns. 
Second, the training objective of recovering first-order missing links does not comply with the high-order graph nature of complex logical queries, and hence KGEs cannot handle complex queries  without training auxiliary logical functions on sampled supervised datasets~\cite{hamilton2018embedding,ren2019query2box,arakelyan2021complex}. 
In addition, existing KG benchmark datasets~\cite{toutanova-chen-2015-observed,xiong2017deeppath,ren2019query2box} usually consist of limited task-specific types of queries, practically prohibiting KGE methods from generalizing well to queries with out-of-domain types.

\vpara{Contributions.}
In this work, we propose to learn deep knowledge representations for answering complex queries from these two perspectives: architecture and training objective. 
At the architecture level, the goal is to design a deep model specified for KGs such that the complex logical queries can be handled. 
At the training level, the model is expected to learn general instead of task-specific knowledge from KGs and is thus enabled with strong generalizability for out-of-domain queries. 
To this end, we present \method---a Transformer-based GNN architecture---with self-supervised pre-training strategies for handling complex logical queries.

\textbf{\method.} 
We develop \method to encode KGs. 
Specifically, to represent relations in KGs, we propose the \textit{Triple Transformation} strategy that turns relations to relation-nodes and thus transforms a KG into a directed graph without edge attributes. 
To further enlarge the model capacity with low computation cost, 
we adopt the \textit{Mixture-of-Experts} strategy to leverage the sparse activation nature of Transformer's feed-forward layers. 
Altogether, these strategies enable the \method architecture with high-capacity and computational efficiency, making it capable of answering the EPFO queries on KGs.

\textbf{Masked Pre-Training.} 
To further improve \method's generalizability, we introduce a masked pre-training framework to train it.  
We formulate complex logical query answering as a masked prediction problem. 
During pre-training, we randomly sample subgraphs from KGs and mask random entities for prediction. 
It includes two sequential stages of \textit{dense initialization}, which targets at enriching the model by training on dense and arbitrary-shaped contexts, and \textit{sparse refinement}, which is trained on sparse and clean meta-graphs to mitigate the gap between pre-training and downstream queries. 


Extensive experiments on two widely-used KG benchmarks, i.e., FB15k-237 and NELL995, demonstrate \method's performance advantages over state-of-the-art---particularly KGE-based---approaches on nine in-domain and out-of-domain downstream reasoning challenges. 
Additionally, the case studies suggest that masked pre-training can endow \method's reasoning with explainability and interpretability via providing predictions over unknown intermediates.

\section{The EPFO Logical Queries}

We introduce the Existential Positive First-Order (EPFO) logical queries on KGs~\cite{ren2019query2box} and identify the unique challenges. 


\vpara{EPFO.}
Let $\mathcal{G}=(\mathcal{E},\mathcal{R})$ denote a KG, where $e\in\mathcal{E}$ denotes an entity and $r\in\mathcal{R}$ is a binary predicate (or relation) $r:\mathcal{E}\times\mathcal{E}\to\rm{\{True, False\}}$ that indicates whether a relation holds for a pair of entities. 
Given the First-Order logical existential ($\exists$) and conjunctive ($\wedge$) operations, the conjunctive queries are defined as:
\beqn{
\begin{aligned}
    \mathcal{Q}[A] \triangleq ?A : \ & \exists E_{1}, \ldots, E_{m}. e_{1} \land \ldots \land e_{n} \\
    \text{where} \ & e_{i} = r(c, E), \text{ with } E \in \{ A, E_{1}, \ldots, E_{m} \}, c \in \mathcal{E}, r \in \mathcal{R} \\
    \text{or} \ & e_{i} = r(E, E^{\prime}), \text{ with } E, E^{\prime} \in \{ A, E_{1}, \ldots, E_{m} \}, E \neq E^{\prime}, r \in \mathcal{R}.
\end{aligned}
}where $A$ refers to the (unknown) target entity of the query, $E_{1}, \ldots, E_{m}$ refer to existentially quantified bound variables (i.e., unknown intermediate entity sets), and $c$ refers to the source entity. 
Given the query $\mathcal{Q}$, the goal is to find its target entity set $\mathcal{A} \subseteq \mathcal{E}$ that satisfies $a \in \mathcal{A}$ iff $\mathcal{Q}[a]$ is true. 

Besides the conjunctive queries, EPFO also covers the disjunctive ($\vee$) queries. 
A rule-of-thumb practice is to transform an EPFO query into the Disjunctive Normal Form~\cite{davey2002introduction,ren2019query2box,arakelyan2021complex}. In other words, a disjunctive query can be decomposed into several conjunctive queries, and a rule can be applied to synthesize conjunctive results for disjunctive predictions. 

\begin{figure*}[t]
    \centering
    \includegraphics[width=.99\textwidth]{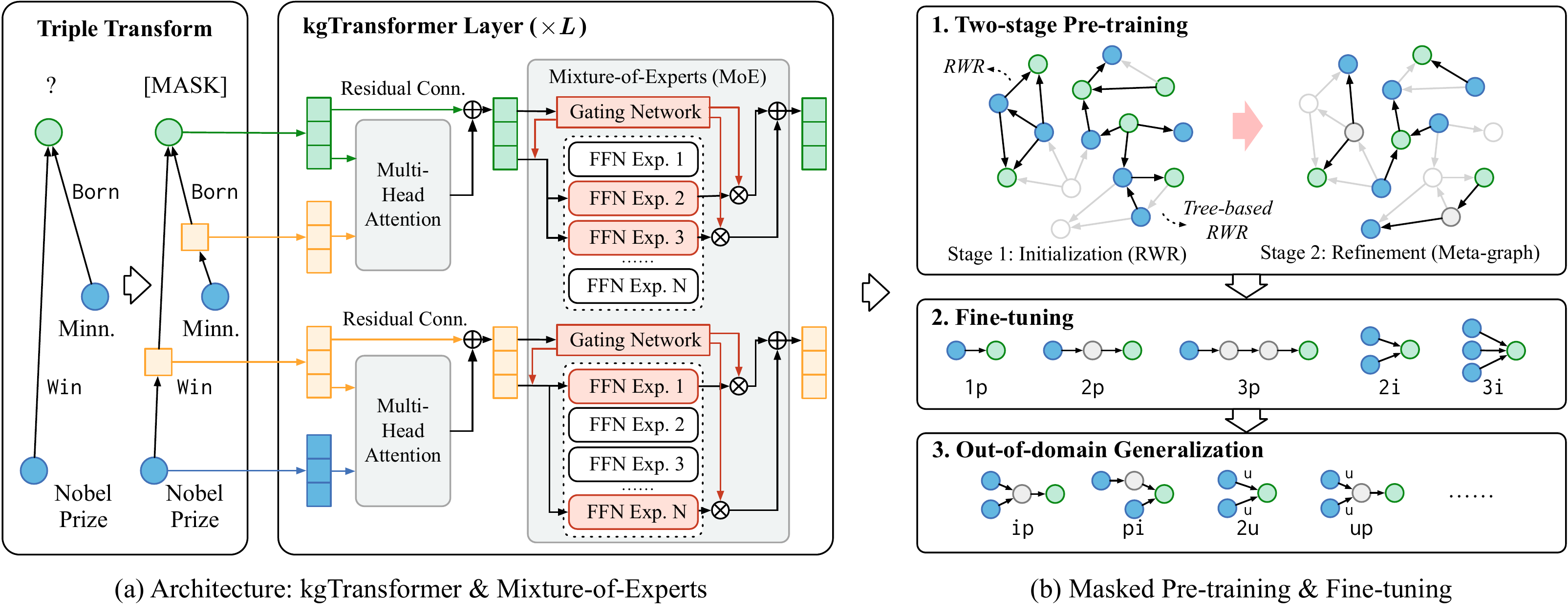}
    \caption{The KG Pre-Training and Reasoning Framework.    
    \textmd{ (a) \method with Mixture-of-Experts is a high-capacity architecture that can capture EPFO queries with exponential complexity.
    (b) Two-stage pre-training trades off general knowledge and task-specific sparse property. 
    Together with fine-tuning, \method can achieve better in-domain performance and out-of-domain generalization.}}
    \label{fig:framework}
\end{figure*}

\vpara{Challenges.}
Compared to the KG completion task in which  the KGE-based methods are prevalent, the EPFO queries can be multi-hop; their numerous combinations the test reasoner's out-of-domain generalizability. 
All these characteristics together pose the following unique challenges to reasoners:

\begin{itemize}[leftmargin=*,itemsep=0pt,parsep=0.2em,topsep=0.3em,partopsep=0.3em]
    \item \textbf{Exponential complexity:} 
    The complexity of EPFO queries grows exponentially as the hop increases~\cite{ren2021smore}, requiring high-capacity and advanced models to handle them. 
    KGE-based reasoners rely on embeddings and simple operators~\cite{ren2019query2box,klement2013triangular} to reason in a ``left-to-right'' autoregressive fashion. 
    However, there are evidences~\cite{arakelyan2021complex} showing that such models' performance gradually saturates as the embedding dimension grows to 1000.
    Additionally, during reasoning, the first-encoded entities are unaware of the later-encoded, ignoring the useful bidirectional interactions.
    
    \item \textbf{Transfer and generalization:} 
    After training, an ideal reasoner is expected to transfer and  generalize to out-of-domain queries. 
    But existing EPFO reasoners~\cite{ren2019query2box,arakelyan2021complex,kotnis2021answering} are directly trained on a limited number of samples within a few query types (i.e., 1p, 2p, 3p, 2i, and 3i in Figure~\ref{fig:framework} (b)) in a supervised manner, leaving many  entities and relations in original KGs untouched. 
  Thus the reasoners are prohibited from grasping knowledge of diverse forms and larger contexts beyond those existing types can express, consequently harming the generalizability.
\end{itemize}

In summary, these challenges make EPFO queries different from conventional KG completion, which only involves single-hop and single-type  queries. 
In this work, we explore how to effectively handle EPFO queries with the pre-training and fine-tuning paradigm. 

\section{The KG Pre-Training Framework}
In this section, we introduce \method---a Transformer-based graph neural network (GNN)---for handling EPFO queries on KGs. 
To endow \method with strong generalization, we design a masked pre-training and fine-tuning framework. 
The overall KG pre-training and reasoning framework is illustrated in Figure ~\ref{fig:framework}. 

\subsection{The \method Architecture}
As discussed above, the relatively simple architecture of KGE-based reasoners limits their expressiveness. 
To overcome this issue, we propose a Transformer-based GNN architecture, \method, with the Mixture-of-Expert strategy to scale up model parameters while keeping its computational efficiency.

\vpara{Transformer for KGs.}
Transformer~\cite{vaswani2017attention}, a neural architecture originally proposed to handle sequences, has achieved early success in the graph domain~\cite{hu2020heterogeneous}. 
To apply transformers to KGs, there are two questions to answer: \textit{1) How to encode the node adjacency in graph structures}, and \textit{2) how to model both entities and relations}.

First, there have been well-trodden practices for encoding node adjacency in graph transformers. 
For node and graph classification, it is common to view graphs as sequences of tokens with positional encoding, ignoring the adjacency matrices~\cite{yun2019graph,ying2021transformers}. 
For link prediction, however, adjacency matrices can be crucial and should be masked to self-attention for better performance~\cite{hu2020heterogeneous}.
As EPFO reasoning is intrinsically a link prediction problem, we follow HGT~\cite{hu2020heterogeneous} to mask adjacency matrices to self-attention.

Second, how to incorporate both entities and relations into Transformer's computation has thus far seldom studied. 
Here, we present the \textit{Triple Transformation} operation, denoted as function $L(\cdot)$, to turn relations into relation-nodes and consequently transform KGs into directed graphs without edge attributes.

\begin{figure}
    \small
    \centering
    \includegraphics[width=.95\linewidth]{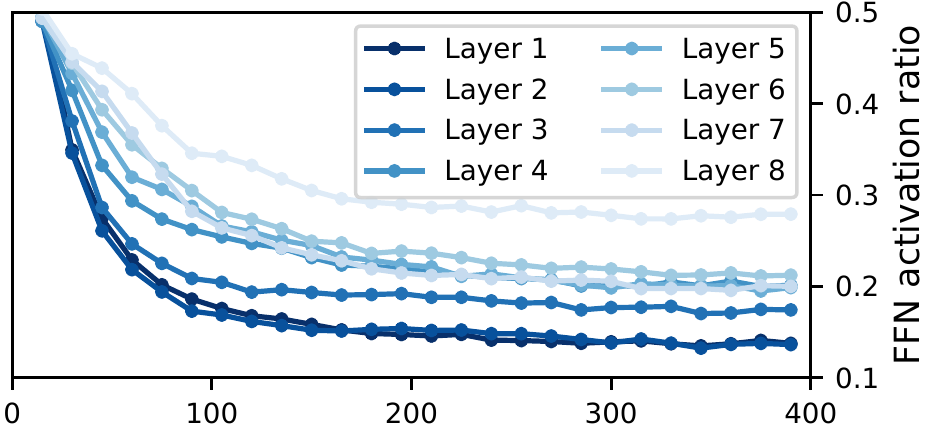}
    \vspace{-3mm}
    \caption{ FFN's activation ratio of \method along pre-training steps (w/o MoE) in preliminary experiments. }
    \label{fig:activation}
    \vspace{-3mm}
\end{figure}

For each directed triple $(h, r, t)$ in $\mathcal{G}$ with $r(h, t)=\text{True}$, we create a node $r_{ht}$ in $L(\mathcal{G})$ (i.e., a \textit{relation-node}) that connects to both $h$ and $r$. 
By mapping each relation edge in $\mathcal{G}$ to a relation-node, 
the resultant graph becomes $L(\mathcal{G})=(\mathcal{E}^\prime, \mathcal{R}^\prime)$ where $\mathcal{E}^\prime=\mathcal{E}\cup\{r_{ht}|r(h, t)=\text{True}\}$ and the unattributed edge set  $\mathcal{R}^\prime=\{r:r(h,r_{ht})=r(r_{ht},t)=\text{True}|r(h,t)=\text{True}\}$. 
In practice, the computational cost of the triple transformation is low as the reasoning graphs for EPFO queries are usually very small and sparse (\textasciitilde{}$10^1$ entities and relations). 

Given the input embeddings $\mathbf{x}_e^{(0)}\in\mathbb{R}^d$ for $e\in\mathcal{E}^\prime$ with dimension $d$, we add a special node-type embedding $t_{\mathbb{I}(e\in\mathcal{E})}\in\mathbb{R}^d$ to distinguish whether it is an entity-node or a relation-node. In the $k$-th layer  of \method, $d_h=d/H$ and $H$ denotes the number of attention heads, and the multi-head attention is computed as
\begin{equation}
    \begin{gathered}
        \text{Attn}_i^{(k)} = \textrm{softmax}(Q^\intercal K/\sqrt{d_h})V, 
    \end{gathered}
\end{equation}
where $Q=\mathbf{x}_e^{(k-1)} \mathbf{W}_Q$ and $\{K, V\}= \bigparallel_{n\in\mathcal{N}(e)} \mathbf{x}_n^{(k-1)} \{\mathbf{W}_K, \mathbf{W}_V\}$. 
Here $\mathbf{W}_{\{Q,K,V\}}\in\mathbb{R}^{d\times d_h}$, $\bigparallel$ denotes concatenation, and $\mathcal{N}(e)$ refers to node $e$'s neighbor set. 

Next, the feed-forward network $\text{FFN}(x)$ is applied to attention-weighted outputs projected by $\mathbf{W}_O\in\mathbb{R}^{d\times d}$ as
\begin{equation}
    \begin{gathered}
        \mathbf{x}_e^{(k)} = \text{FFN}(\bigparallel_{i=1}^H \text{Attn}_i^{(k)}\cdot\mathbf{W}_O), \text{FFN}(\mathbf{x}) = \sigma(\mathbf{x}\mathbf{W}_1 + \mathbf{b}_1)\mathbf{W}_2 + \mathbf{b}_2
    \end{gathered}
\end{equation}
where $\mathbf{W}_1\in\mathbb{R}^{d\times 4d}$, $\mathbf{W}_2\in\mathbb{R}^{4d\times d}$, and $\sigma$ is the activation function (e.g., GeLU~\cite{hendrycks2016gaussian}). 
Note that FFN is critical for Transformers to capture massive patterns and proved equivalent to the key-value networks~\cite{geva2021transformer}.

Different from 1) existing KGE-based reasoners with the ``left-to-right'' reasoning order and 2) the sequence encoder-based model~\cite{kotnis2021answering} that can only reason on acyclic query graphs, \method designs an architecture to aggregate information from all directions in each layer's computation, making it more flexible and capable of  answering queries in arbitrary shapes. 
Figure ~\ref{fig:framework} (a) illustrates the \method architecture.

\vpara{Mixture-of-Experts (MoE).}
Though \method's architecture allows it to capture complicated reasoning patterns, the number of its parameters soars up quadratically with the embedding dimension $d$, which is a common challenge faced by Transformers. 


As mentioned earlier, Transformer's FFN is known to be equivalent to the key-value networks~\cite{geva2021transformer} where a key activates a few values as responses. 
Given $x\in\mathbb{R}^d$, FFN's intermediate activation 
\begin{equation} \label{eq:activation}
    \begin{aligned}
        \sigma(\mathbf{x}\mathbf{W}_1+\mathbf{b}_1) = [\mathbf{x}_0, \mathbf{x}_1, ..., \mathbf{x}_i, ..., \mathbf{x}_j, ..., \mathbf{x}_{4d}] = \underbrace{[0, 0, ..., \mathbf{x}_i, ..., \mathbf{x}_j, ..., 0]}_{\text{Most of elements are 0}}
    \end{aligned}
\end{equation}
\noindent can be extremely sparse, where $\mathbf{W}_1\in\mathbb{R}^{d\times 4d}$. 
The level of sparsity varies with tasks. 
For instance,  a recent study ~\cite{zhang2021moefication} shows that usually less than 5\% of neurons are activated for each input in NLP. 
In our preliminary experiments on EPFO queries (Cf. Figure~\ref{fig:activation}), only 10\%-20\% neurons are activated for certain inputs (except the last decoder layer).

Thus, we propose to leverage the sparsity of \method via the Mixture-of-Experts (MoE) strategy~\cite{rokach2010pattern,shazeer2017outrageously}. 
MoE first decomposes a large FFN into blockwise experts, and then utilizes a light gating network to select experts to be involved in the computation. 
For example, an FFN with $\mathbf{W}_1\in\mathbb{R}^{d\times16d}$ and $\mathbf{W}_2\in\mathbb{R}^{16d\times d}$ (4 times larger than that in Equation~\ref{eq:activation}) can be transformed into
\begin{equation}
    \begin{aligned}
    \text{FFN}(\mathbf{x}) = \{\text{FFN}_{\rm Exp}^{(i)}(\mathbf{x})=\sigma(\mathbf{x}\mathbf{W}_1^{(i)} + \mathbf{b}_1)\mathbf{W}_2^{(i)} + \mathbf{b}_2~|~i\in[1, 8], i\in\mathbb{N}\}
    \end{aligned}
\end{equation}
\noindent where $\mathbf{W}_1^{(i)}\in\mathbb{R}^{d\times2d}$, $\mathbf{W}_2^{(i)}\in\mathbb{R}^{2d\times d}$. 
Each $\text{FFN}_{\rm Exp}^{(i)}$ is referred to as an \textit{expert}. 
Given a gating network, which is usually a trainable matrix $\mathbf{W}_{\rm gate}\in\mathbb{R}^{d\times N}$ with $N$ as the number of experts,  we can select the top-2 experts for computation ~\cite{lepikhin2020gshard} as
\begin{equation}
    \begin{aligned}
    \mathbf{x}_e^{(k)} = \sum_{(i, s)\in S}s\cdot\text{FFN}_{\rm Exp}^{(i)}(\mathbf{x}_e^{(k-1)})
    \end{aligned}
\end{equation}
\noindent where $S=\{(i, s)| s\in\text{softmax}(\text{Top-2}(\mathbf{x}_e^{(k-1)}\mathbf{W}_{\rm gate})), i\text{ is the index of } s\}$. 
By using MoE, we can significantly enlarge the actual model size with approximately the same training computation as the FFN in Equation~\ref{eq:activation}. 
During inference, all experts are included in the computation and the FFN output becomes the weighted average of every expert's output.
The implementation details can be found in Appendix~\ref{app:Reproducibility}.

\subsection{Masked Pre-Training and Fine-Tuning}
To further improve the generalizability of \method, we present a masked pre-training and fine-tuning framework for KG reasoning, which is illustrated in Figure~\ref{fig:framework} (b). 

\subsubsection{Two-stage Pre-training: Initialization and Refinement}~\\
Most existing EPFO reasoners train  the embeddings over a limited number of sampled queries with few specific query types in a supervised manner~\cite{ren2019query2box}. 
The low coverage of entities and relations in original KGs in training usually leads to poor transferability to queries with unseen entities in testing.
In addition, the reasoners are expected to answer out-of-domain types of queries after training. 
For example, the standard benchmark~\cite{ren2019query2box} trains the model on 5 types of queries (1p, 2p, 3p, 2i, and 3i), and asks the model to test on additional types of queries (ip, pi, 2u, and up). 


To overcome this challenge, we propose to do masked pre-training for \method. 
The main idea is to randomly sample arbitrary-shaped masked subgraphs from the original KGs' training set to pre-train the model. 
Its advantages lie  in the large receptive fields of sampled subgraphs, diverse shapes of queries, and the high coverage over original KGs. 
To fully explore general knowledge from pre-training, we should sample masked query graphs as dense and large as possible; however, the query graphs during downstream reasoning are usually small and sparse. 
Thus, such strategy would cause a mismatch between pre-training and downstream distributions, resulting in performance degradation. 



To mitigate the issue, we introduce a \textit{two-stage} masked pre-training strategy for \method. 
The first stage aims to initialize \method with KGs' general knowledge, and the second one further refines its ability for small and sparse queries during inference.

\vpara{Stage 1: Dense Initialization.} 
The goal of this stage is to enable \method with the general knowledge in KGs via masked pre-training over dense and large sampled subgraphs. 

We use two random-walk-based strategies to sample on original KGs' training set. 
One is the random walk with restart (RWR), for which we make no assumption on the shapes of subgraphs it may sample. 
Another is a tree-based RWR strategy, for which we constrain the shape of sampled graphs as tree structures to cater conjunctive queries. 
We include a majority of induced relations between sampled entities for denser contexts. 

Both methods are used to produce sampled query subgraphs. 
Given a \method parameterized by $\theta$, a set of random sampled entities $\mathcal{E}_{\rm mask}$ that are masked by a special mask embedding turn the trainable input embeddings $\mathbf{X}=[\mathbf{x}_{e_0}^{(0)}, ..., \mathbf{x}_{e_T}^{(0)}]$ into mask corrupted $\mathbf{\hat{X}}$. 
The dense initialization stage seeks to optimize 
\begin{equation}
    \begin{aligned}
        \mathcal{L}_{\rm init} = - \sum_{e_{\rm m}\in\mathcal{E}_{\rm mask}} \mathop{\log} p_\theta(\mathbf{x}_{e_{\rm m}} | \mathbf{\hat{X}}, \mathbf{A})
    \end{aligned}
\end{equation}
where $\mathbf{A}$ is the adjacency matrix. 
The prediction is done simultaneously for all masked nodes, rather in an autoregressive manner.

\vpara{Stage 2: Sparse Refinement.} 
The goal of this stage is to further enhance the model's  capacity by using meta-graph sampling for the EPFO queries that are commonly sparse and small. 

We adopt the basic EPFO patterns (1p, 2p, 3p, 2i, and 3i) as meta-graphs, which involve no more than four entities and relations. 
No induced relations between entities are reserved. 
Following real query practices, except for source entities (``\condent{}'' in Figure~\ref{fig:framework}(b)), other entities are all masked but only the target entity (``\maskent{}'') is predicted in each query type. 

Given each meta-graph type, the set of masked entities $\mathcal{E}_{\rm mask}=\mathcal{E}_{\rm inter}\cup\mathcal{E}_{\rm target}$ where $\mathcal{E}_{\rm inter}$ and $\mathcal{E}_{\rm target}$ refer to the sets of intermediate and target entities, respectively. 
The sparse refinement stage only optimizes the error $\mathcal{L}_{\rm refine}$ on target entity $e_{\rm t}\in\mathcal{E}_{\rm target}$ as the same form of $\mathcal{L}_{\rm init}$ as 
\begin{equation}
    \begin{aligned}
        \mathcal{L}_{\rm refine} = - \sum_{e_{\rm t}\in\mathcal{E}_{\rm target}} \mathop{\log} p_\theta(\mathbf{x}_{e_{\rm t}} | \mathbf{\hat{X}}, \mathbf{A})
    \end{aligned}
\end{equation}
During \method pre-training, these two stages are conducted sequentially rather than in parallel.

\subsubsection{Fine-tuning}~\\\label{arch:finetune}
We introduce the fine-tuning strategy of \method for downstream reasoning tasks. 
Though with the sparse refinement stage during pre-training, our preliminary experiments (Cf. Table~\ref{tab:ablation}) show that fine-tuning is still necessary for \method as the labeled supervision information is used to tune the pre-trained model for downstream tasks.

Specifically, we fine-tune the pre-trained \method using the downstream EPFO reasoning datasets in the form of masked prediction. 
Given a set of feasible answers $\mathcal{E}_A$ for a masked query graph, the fine-tuning loss function per query is formulated as
\begin{equation}
    \begin{aligned}
        \mathcal{L}_{\rm ft} = -\frac{1}{|\mathcal{E}_A|}\sum_{e_a\in\mathcal{E}_A}\mathop{\log}\frac{\mathop{\exp}(\mathbf{x}_{e_a}^{(L)\intercal} \mathbf{u}_{e_a})}{\mathop{\exp}(\mathbf{x}_{e_a}^{(L)\intercal} \mathbf{u}_{e_a}) + \sum_{e\notin\mathcal{E}_A}\mathop{\exp}(\mathbf{x}_e^{(L)\intercal}\mathbf{u}_e)}
    \end{aligned}
\end{equation}
where other answers' predicted logits are ignored when computing the loss for one answer, and per query loss is averaged among all answers' losses. 
$\mathbf{u}_e$ refers to entity $e$'s embedding in the decoder.

In practice, an intuitive strategy is to fine-tune the pre-trained \method on the corresponding downstream training set for each reasoning task. 
However, the recent NLP studies~\cite{raffel2020exploring} have demonstrated that the multi-task fine-tuning can avoid over-fitting and let tasks benefit from each other. 
Inspired from this observation, we first jointly fine-tune \method on all possible downstream training sets, and then fine-tune it for each single task. 
In addition, we  observe that a combination of several downstream training sets can sometimes be much better than one single task's training set (Cf. Table~\ref{tab:multi-task}). 
Finally, the best checkpoint for testing is selected on the basis of per task validation set performance.


\subsubsection{Out-of-domain Generalization}~\\
To better test the model's generalizability to unseen queries, four specific types of out-of-domain queries, \emph{ip, pi, 2u, up}, are provided only in the validation and test sets as illustrated in Figure~\ref{fig:framework} (b). 

Among them, the conjunctive queries \emph{ip} and \emph{pi} can be represented as query graphs. 
For the disjunctive/union queries \emph{2u} and \emph{up}, we adopt the idea of Disjunctive Normal Form~\cite{davey2002introduction}. 
Specifically, we first predict on the decomposed conjunctive queries respectively, then normalize each probability distribution into its rank, and finally combine the ranks by taking the highest rank. 
Instead of taking the mean or max of probabilities, we find that re-scoring the entity prediction by its rank from the probability distribution in each decomposed conjunctive query is more effective, as the  probability distributions for different decomposed queries can be of different scales.


\begin{table*}[t]
	\centering
    \caption{Hits@3m for complex query reasoning. \textmd{(\textbf{bold} denotes the best results; \underline{underline} denotes the second best results)}.}
    \renewcommand\arraystretch{1.1}
    \renewcommand\tabcolsep{5pt}
    \begin{threeparttable}
\begin{tabular}{@{}ll|p{0.8cm}<{\centering}p{0.8cm}<{\centering}|p{0.8cm}<{\centering}p{0.8cm}<{\centering}p{0.8cm}<{\centering}p{0.8cm}<{\centering}p{0.8cm}<{\centering}p{0.8cm}<{\centering}p{0.8cm}<{\centering}p{0.8cm}<{\centering}p{0.8cm}<{\centering}@{}}
\toprule[1.2pt]
\multirow{2}{*}{Dataset} & \multirow{2}{*}{Model}    & \multirow{2}{*}{Avg}                      & \multirow{2}{* }{\makecell[c]{Avg\\w/o u}}                               & \multicolumn{5}{c}{In-domain}                                                                                                                                                      & \multicolumn{4}{c}{Out-of-domain}                \\ \cmidrule(l){5-9} \cmidrule(l){10-13}
\multicolumn{1}{l}{} & \multicolumn{1}{l}{}    & \multicolumn{1}{|l}{} &  & \multicolumn{1}{c}{1p} & \multicolumn{1}{c}{2p} & \multicolumn{1}{c}{3p} & \multicolumn{1}{c}{2i} & \multicolumn{1}{c}{3i} & \multicolumn{1}{c}{ip} & \multicolumn{1}{c}{pi} & \multicolumn{1}{c}{2u} & \multicolumn{1}{c}{up} \\ \midrule
\multirow{8}{*}{NELL995}& \multicolumn{1}{|l|}{GQE~\cite{hamilton2018embedding}}                                    & 0.248                                   & 0.270                                                & 0.417                   & 0.231                   & 0.203                   & 0.318                   & 0.454                   & 0.081                   & 0.188                   & 0.200                   & 0.139                  \\
                          & \multicolumn{1}{|l|}{Q2B~\cite{ren2019query2box}}                                       & 0.306                                      & 0.317                                                & 0.555                   & 0.266                   & 0.233                   & 0.343                   & 0.480                   & 0.132                   & 0.212                   & 0.369                   & 0.163                  \\
                          & \multicolumn{1}{|l|}{EmQL~\cite{sun2021faithful}$^1$}                                   & 0.277                                      & 0.294                                                & 0.456                   & 0.231                   & 0.172                   & 0.331                   & 0.483                   & 0.143                   & 0.244                   & 0.226                   & 0.207                  \\
                          & \multicolumn{1}{|l|}{BiQE~\cite{kotnis2021answering}}                                   & \multicolumn{1}{c}{-}                      & 0.344                                                & 0.587                      & 0.305                   & \underline{0.326}                   & 0.371                   & \underline{0.531}                   & 0.103                   & 0.187                   & \multicolumn{1}{c}{-}   & \multicolumn{1}{c}{-}  \\
                          & \multicolumn{1}{|l|}{CQD\footnotesize{(CO)}\normalsize~\cite{arakelyan2021complex}}     & 0.368                                      & 0.370                                                & \bf 0.667                      & 0.265                   & 0.220                   & \bf 0.410                   & 0.529                   & \underline{0.196}                   & \underline{0.302}                   & \bf 0.531                   & \underline{0.194}                  \\
                          & \multicolumn{1}{|l|}{CQD\footnotesize{(Beam)}\normalsize~\cite{arakelyan2021complex}}   & \underline{0.375}                                      & \underline{0.385}                                                & \bf 0.667                   & \underline{0.350}                   & 0.288                   & \bf 0.410                   & 0.529                   & 0.171                   & 0.277                   & \bf 0.531                   & 0.156                  \\ \cmidrule{2-13}
                          & \multicolumn{1}{|l|}{\method} & \bf 0.399                                      & \bf 0.408                                               & \underline{0.625}                   & \bf 0.401                   & \bf 0.367                   & \underline{0.405}                   & \bf 0.546                   & \bf 0.203                   & \bf 0.306                  & \underline{0.469}                  & \bf 0.270                  \\ \midrule
\multirow{8}{*}{FB15k-237}& \multicolumn{1}{|l|}{GQE~\cite{hamilton2018embedding}}                                  & 0.230                                      & 0.250                                                & 0.405                   & 0.213                   & 0.153                   & 0.298                   & 0.411                   & 0.085                   & 0.182                   & 0.167                   & 0.160                  \\
                          & \multicolumn{1}{|l|}{Q2B~\cite{ren2019query2box}}                                       & 0.268                                      & 0.283                                                & 0.467                   & 0.240                   & 0.186                   & 0.324                   & 0.453                   & 0.108                   & 0.205                   & 0.239                   & \underline{0.193}                  \\
                          & \multicolumn{1}{|l|}{EmQL~\cite{sun2021faithful}$^1$}                                   & 0.219                                 & 0.241                                                & 0.389                           & 0.201                   & 0.154                   & 0.275                   & 0.386                   & 0.101                   & 0.184                           & 0.115                   & 0.165                  \\
                          & \multicolumn{1}{|l|}{BiQE~\cite{kotnis2021answering}}                                   & -                                          & 0.293                                                & 0.439                   & 0.281                   & \underline{0.239}                   & 0.333                   & \underline{0.474}                   & 0.110                   & 0.177                   & -                       & -                      \\
                          & \multicolumn{1}{|l|}{CQD\footnotesize{(CO)}\normalsize~\cite{arakelyan2021complex}}     & 0.272                             & 0.290                                                & 0.512                   & 0.213                   & 0.131                   & \underline{0.352}                   & 0.457                   & \underline{0.146}                   & 0.222                   & 0.281                   & 0.132                  \\
                          & \multicolumn{1}{|l|}{CQD\footnotesize{(Beam)}\normalsize~\cite{arakelyan2021complex}}   & \underline{0.290}                             & \underline{0.315}                                                & \bf 0.512                   & \underline{0.288}                   & 0.221                   & \underline{0.352}                   & 0.457                   & 0.129                   & \underline{0.249}                   & \bf0.284                   & 0.121                  \\ \cmidrule{2-13}
                          & \multicolumn{1}{|l|}{\method} & \bf 0.325                                     & \bf 0.350                                                & \underline{0.459}                   & \bf 0.312                   & \bf 0.276                   & \bf 0.398                   & \bf 0.528                   & \bf 0.189                   & \bf 0.286                   & \underline{0.263}                   & \bf 0.214                                 \\  
\bottomrule[1.2pt]
\end{tabular}

    \begin{tablenotes}
        \footnotesize
        \item[1] EmQL's reported results are not under the standard metric. We have verified the mismatch with its authors and re-evaluated the performance.
    \end{tablenotes}
    \end{threeparttable}
    \label{tab:main}
\end{table*}

\begin{table}[t]
    \centering
    \renewcommand\tabcolsep{3pt}
    \caption{Ablation on pre-training \& fine-tuning   (Hits@3m).}
    \begin{tabular}{lcc}
        \toprule[1.2pt]
        & FB15k-237 & NELL995\\ \midrule
        \method (Stage 1 + Stage 2) & \textbf{0.336} & 0.395 \\ \midrule
        \quad -only Stage 1 in pre-training & 0.308 & 0.307\\
        \quad -only Stage 2 in pre-training & 0.307 & \textbf{0.399}\\ \midrule
        \quad -w/o fine-tuning & 0.301 & 0.368\\
        \quad -w/o pre-training & 0.262 & 0.288\\
        \bottomrule[1.2pt]
    \end{tabular}
    \label{tab:ablation}
\end{table}

\section{Experiments}
\label{sec:exp}
In this section, we evaluate \method's capacity to reason for EPFO queries with at least one imputed edge, i.e., for answers that cannot be obtained by direct KG traverses~\cite{ren2020beta}. 
We employ two   benchmarks FB15k-237 and NELL995 with nine different reasoning challenges including both in-domain and out-of-domain queries following prior settings in Query2Box~\cite{ren2019query2box}. 

\vpara{Datasets.}
\label{exp:dataset}
The statistics of FB15k-237 \cite{toutanova-chen-2015-observed} and NELL995 \cite{xiong-etal-2017-deeppath}  can be found in Table~\ref{tab:dataset}.
We use the standard training/validation/test edge splits ~\cite{ren2019query2box} to pre-train the \method model. 
$\mathcal{G}_{train}$ contains the training edges, and the subgraphs are sampled from $\mathcal{G}_{{train}}$ for pre-training. 
We search for hyper-parameters over $\mathcal{G}_{valid}$. 
Here we do not include FB15k as it is known to suffer from major test leakage through inverse relations as illustrated in \cite{toutanova-chen-2015-observed}, which consequently proposes the updated dataset FB15k-237. 
See Table~\ref{tab:split} in Appendix for more detailed information about the datasets. 


In addition, the query-answer datasets $\llbracket q \rrbracket_{\text{train}},\llbracket q \rrbracket_{\text{valid}},\llbracket q \rrbracket_{\text{test}}$ are constructed for logic query reasoning by Query2Box \cite{ren2019query2box}, including chain-shaped queries (\textit{1p, 2p, 3p}), conjunctive queries (\textit{2i, 3i, ip, pi}), and disjunctive queries (\textit{2u, up}). 
For a query $q$, the three sets of answers are obtained by performing subgraph matching over three graphs: $\llbracket q \rrbracket_{train}$ from $\mathcal{G}_{train}$, $\llbracket q \rrbracket_{valid}$ from $\mathcal{G}_{valid}\backslash \mathcal{G}_{train}$ and $\llbracket q \rrbracket_{test}$ from $\mathcal{G}_{test}\backslash \mathcal{G}_{valid}$. 
Therefore, it does not spoil the train/dev/test split in the original dataset.  
We use $\llbracket q \rrbracket_{train}$ for fine-tuning and training, $\llbracket q \rrbracket_{valid}$ for validating, and $\llbracket q \rrbracket_{test}$ for testing. 
Such design allows us to test if the model can predict non-trivial answers that are unable to obtain by traversing the given KG. 

\vpara{The Evaluation Protocol.}
We follow the evaluation protocol in Query2Box~\cite{ren2019query2box} including the filtering setting~\cite{bordes2013translating} and metrics calculation. 
In filtering setting, since the answers to most queries are not unique, we rule out other correct answers when calculating the rank of one answer $a$. 
Hits at K(Hits@Km) is used as the metric, which is different from Hits@K as it requires to first average on Hits@K per query and then average over all queries. 
See more details in Appendix~\ref{app:eval}).

\subsection{Main Results}
We compare \method with various methods based on both KGEs and sequence encoders, including GQE~\cite{hamilton2018embedding}, Q2B~\cite{ren2019query2box}, BiQE~\cite{kotnis2021answering}, EmQL~\cite{sun2021faithful}, and the state-of-the-art baseline CQD~\cite{arakelyan2021complex}. 
Detail descriptions can be found in  Appendix~\ref{app:baseline}.

Table~\ref{tab:main} reports the Hits@3m results for all query types on FB15k-237 and NELL995. 
For FB15K-237, 
Note that the two-stage pre-training is sequentially adopted for FB15K-237 and only the second stage of pre-training is used for NELL995 (Cf. Section \ref{subsec:ablation} for detailed analysis). 
Among most cases, \method obtains the best performance on both datasets. 
In contrast with BiQE, which is also based on Transformer, the \method model achieves average improvements of $6.4\%$ (18.6\% relative) and $5.7\%$ ($19.5\%$ relative)  without union operation on NELL995 and  FB15k-237, respectively. 
It demonstrates the proposed method's architecture superiority in terms of the model capacity and generalizability.
In addition, \method can perform union operations, supporting the complete set of EPFO queries.

Compared to the previous state-of-the-art CQD, \method obtains $2.4\%$ ($6.4\%$ relative) and $3.5\%$ ($12.1\%$ relative) improvements on average over NELL995 and FB15k-237, respectively. 
For out-of-domain reasoning queries on FB15k-237, \method outperforms CQD by significant margins for most types, demonstrating pre-training's capability in helping the model gain out-of-domain generalizability.

The major deficiency of \method happens for the \textit{1p} query, which is identical to the conventional knowledge graph completion problem. 
First, \method's multi-layer Transformer architecture may be unfriendly to queries with limited contexts. 
In addition, the training objective of \method focuses on complex queries with multiple entities and relations instead of those with two entities and one relation in a triplet. 
We will the improvement of \method for the \textit{1p} query for future research.

\begin{figure*}[t]
    \centering
    \includegraphics[width=\textwidth]{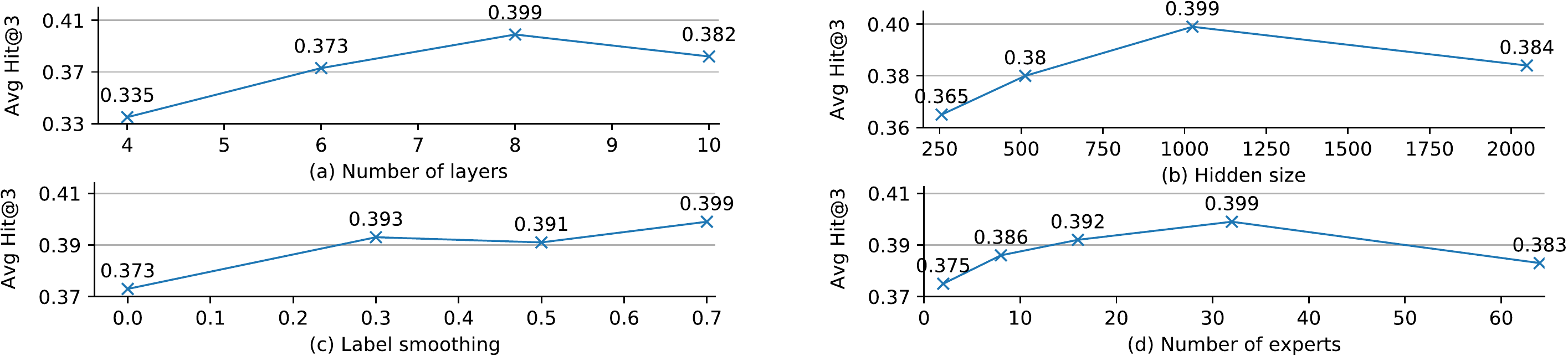}
    \caption{Hyperparameter analysis on NELL995 (Hits@3m). \textmd{The default setting is: 8 layers, 1024 for hidden size, $\alpha=0.7$ for label smoothing, and 32  experts.} }
    \label{fig:ablation}
\end{figure*}

\subsection{Ablation Study}\label{subsec:ablation}
\vpara{Pre-Training and Fine-Tuning Strategies.}
We analyze the necessity of two-stage pre-training and the function of fine-tuning through ablation study on both datasets. 
The contributions of fine-tuning and the pre-training stages are summarized in Table~\ref{tab:ablation}, where the average Hits@3m results for all query types are reported.



We observe that without fine-tuning, the performance of \method drops about 3\% in absolute numbers for both datasets, and further without pre-training, the performance drops are enlarged to 7\% on FB15k-237 and 11\% on NELL995. 
This demonstrates that both pre-training and fine-tuning can help improve the model capacity of \method for complex query reasoning. 
In addition, the pre-trained \method without fine-tuning can achieve comparable (0.368 vs. 0.375 on NELL99) or even better (0.301 vs. 0.290 on FB15k-237) results than CQD~\cite{arakelyan2021complex}---the previously state-of-the-art method.  

Furthermore, two pre-training stages introduce different inductive biases towards the underlying logic. 
\method with only stage 1 pre-training assumes the dense correlation between pairs of entities, while its stage 2 pre-training assumes that the relations among entities follow the chain-rule. 
We observe that \method with both stages obtains the good results, while \method with only stage 2 pre-training generates similar performance on NELL995, due to the fact that it is sparser than FB15k-237 with weaker relations within clusters and stronger connections in chains.

\vpara{MoE Efficiency.}
We present an efficiency analysis in terms of the numbers of experts in pre-training, fine-tuning, and inference stages 
with a batch size of 64 in both training and inference. 
Table~\ref{tab:moe_speed} reports the running time in {millisecond}.
We observe that MoE can significantly increase the model capacity with limited increase of computational cost. 
Take the case of 32 experts for example, MoE helps to enlarge \method 16$\times$ in size by only taking 11.6\%-38.7\% of additional computation.


\begin{table}[t]
    \centering
    \caption{Training and inference time (ms) per step  along with different numbers of experts using per step batch size 64.}
    \renewcommand\arraystretch{1.1}
    \renewcommand\tabcolsep{7pt}
    \begin{tabular}{@{}lcccc@{}}
    \toprule[1.2pt]
     \multirow{2}{*}{} & \multicolumn{4}{c}{Number of experts}\\\cmidrule(l){2-5}
     \multicolumn{1}{l}{} & 2 & 8 & 16 & 32  \\ 
     \midrule
     Pre-training stage 1 &41.46 & 42.02 & 45.38 & 49.30 \footnotesize(+18.9\%) \\
     Pre-training stage 2 &17.37 & 18.83 & 19.78 & 24.09 \footnotesize(+38.7\%) \\
     Fine-tuning & 32.41 & 32.91 & 36.46 & 37.47 \footnotesize(+15.6\%)  \\
     Inference & 12.86 &  13.36 & 13.85 & 14.35 \footnotesize(+11.6\%)  \\
     \bottomrule[1.2pt]
\end{tabular}
    \label{tab:moe_speed}
    \vspace{-3mm}
\end{table}

\begin{table*}[t]
	\centering
    \caption{Case study on 2p examples.
    \textmd{\textbf{(a) Interpretability.}\label{tab:interpret} Filling a certain entity as the tail prediction, we use \method to predict the unknown intermediates to test its interpretability. \textbf{(b) All-direction Reasoning.} In 2p, intermediate and tail entities are masked.
    \method considers both relations when predicting intermediate node (marked \textbf{\textcolor{orange}{orange}}). If the second relation is masked (i.e., conventional autoregressive reasoning), intermediate's predictions are valid for the first relation, but no longer for the second (marked \textcolor{red}{red}). }}
    \renewcommand\tabcolsep{10pt}
    \renewcommand\arraystretch{1.25}
    
\begin{tabular}{cccccc}
\toprule[1.2pt]
\textbf{Case} & \textbf{Head}&&\textbf{Intermediate entity}&&\textbf{Tail}\\
&(Groundtruth)&&(Top 3 prediction)&&(Prediction)\\
\hline
Interpretability&{Elmira(U.S. City)}&{\textit{location in}}&{U.S.}, {Canada}, {UK}&\textit{in country}&Davidson College\\ 
\midrule
\multirow{2}{*}{\makecell[c]{All-direction\\Reasoning}}&\multirow{2}{*}{\textit{Salt}(2010 film)}&\multirow{2}{*}{\textit{has subject}}&\textbf{\textcolor{orange}{CIA}}, Espionage, PTSD&\textit{employs}&George Bush\\ \cmidrule{4-4} \cmidrule{6-6}
&&&\textcolor{red}{Serial killer}, Espionage, PTSD&\textit{(masked)}&-\\
\bottomrule[1.2pt]
\end{tabular}

    \label{tab:2p_case}
\end{table*}

\vpara{Hyperparameters.}
We conduct the ablation experiments on hyperparameters, including the number of layers, hidden size, the value of label smoothing, and the number of experts. 
Figure \ref{fig:ablation} reports the results on NELL995 in terms of Hits@3m. 

\begin{itemize}[leftmargin=*,itemsep=0pt,parsep=0.2em,topsep=0.3em,partopsep=0.3em]
\item \textbf{Number of layers:} 
In Figure~\ref{fig:ablation} (a), we test the performance of different number of layers.
It suggests that a sufficient and proper model depth (8 in this case)  is essential for \method, which can be attributed to the relation between the depth and the size of the receptive field. 
Previously studies have shown that very deep GNNs may suffer from over-smoothing and training instability, thus causing performance drops~\cite{feng2020grand}.

\item \textbf{Hidden size:} 
Figure~\ref{fig:ablation} (b) studies the influence of hidden size. 
We observe that the increase of hidden size helps to capture massive EPFO reasoning patterns, but a too-large one (e.g., 2048) can harm the performance, as it may suffer from the similar training difficulty witnessed in KGE-based reasoners~\cite{arakelyan2021complex}.

\item \textbf{Label smoothing:} 
Label smoothing is crucial to \method's pre-training, as it mitigates the bias in the pre-training data due to the on-the-fly sampling. 
Through experiments in Figure~\ref{fig:ablation} (c), the optimal value $\alpha=0.7$ for NELL995 is exceptionally high in comparison with that in natural language processing ($\alpha=0.1$). 
It indicates that by training for long epochs (around 1000 epochs) with a rather large embedding size (1024), the \method model is prone to overfitting in pre-training.  
We also observe that exclusion of label smoothing will yield a considerable performance decrease by $2.6\%$. 

\item \textbf{Number of experts:} 
We evaluate the performance of \method with different numbers of experts in Figure~\ref{fig:ablation} (d).  
The correspondence of sparse nature in KG reasoning and mixture-of-experts is one key point in \method. 
Compared with the vanilla Transformer (2 experts), \method achieves performance improvements with more and more experts until 32 (16$\times$ large in model size), demonstrating the power of  high-capacity models in EPFO challenges. 

\end{itemize}

\subsection{Case Study}

\vpara{Interpretability.}
In masked querying, the model is enforced to produce plausible predictions for {every} node in the graph because it does not know the queried nodes in advance. 
Such property implies the interpretability of the results. 
The embeddings of intermediate variables can be fed into the prediction layer to generate a result, providing paths to understand how the model produces the answers. 

Take the query ``Which school is in the same country as Elmira?''---formally, $?A, \exists E, \text{location in}(Elmira, E)\land \text{in country}(E, A)$---in Table \ref{tab:2p_case} for example, the predicted answer is ``Davidson College''. 
To verify this answer, we check what the result of $E$ is. 
Inputting the embedding of $E$ into the decoder, it actually gives ``U.S.'' as the top prediction, which is indeed a valid intermediate entity. 
More cases concerning interpretability for 3p queries are provided in Table~\ref{tab:3p_case} in Appendix.

\vpara{All-Direction Reasoning.}
Unlike almost all reasoners' step-by-step searching, \method is a Transformer-based GNN that captures information from all directions simultaneously. 
As the second row in Table \ref{tab:2p_case} shows that when the intermediate entity is required to fit the \textit{employs} relation, ``CIA'' is more suitable than ``Espionage'' and ``PTSD'' as it is an agency. 
Without the ability to read from tail to head, the model would suggest ``Serial killer'' instead, which is inadequate for future deductions.


\tohide{

\begin{table*}[t]
	\centering
    \caption{Main Hits@3m for complex query reasoning on FB15k-237 and NELL995 benchmarks \textmd{(\textbf{bold} denotes the best results; \underline{underline} denotes the second best results)}}
    \renewcommand\arraystretch{1.1}
    \renewcommand\tabcolsep{5pt}
    \begin{threeparttable}
    
    \begin{tablenotes}
        \footnotesize
        \item[1] EmQL's reported results are not under the standard metric. We have verified the mismatch with its authors and re-evaluated the performance.
    \end{tablenotes}
    \end{threeparttable}
    \label{tab:main}
\end{table*}

\section{Experiments}
\label{sec:exp}
In this section, we present extensive evaluations on \method's ability to reason for EPFO queries with at least one edge to be imputed. In other words, we aim at non-trivial answers that cannot be obtained by directly traversing the KG~\cite{ren2020beta}. We introduce two well-established benchmarks FB15k-237 and NELL995 and nine different reasoning challenges including both in-domain and out-of-domain queries following prior settings in Query2Box~\cite{ren2019query2box}. 

\vpara{Datasets.}
\label{exp:dataset}
We evaluate our model on two standard KG datasets: FB15k-237 \cite{toutanova-chen-2015-observed} and NELL995 \cite{xiong-etal-2017-deeppath}, whose statistics can be found in Table~\ref{tab:dataset}.
We use their standard training/validation/test edge splits to pre-train our model following~\cite{ren2019query2box}. $\mathcal{G}_{train}$ contains the training edges and subgraphs are sampled from $\mathcal{G}_{\text{train}}$ for pre-training. We search for hyper-parameters over $\mathcal{G}_{valid}$. See Appendix~\ref{tab:split} for more detailed statistics of datasets. Here we do not include FB15k as it is known to suffer from major test leakage through inverse relations as illustrated in \cite{toutanova-chen-2015-observed}, which consequently proposes a cleaner and better version FB15k-237. 

Besides, for logic query reasoning, query-answer datasets $\llbracket q \rrbracket_{\text{train}},\llbracket q \rrbracket_{\text{valid}},\llbracket q \rrbracket_{\text{test}}$ are constructed by Query2Box \cite{ren2019query2box} including chain-shaped queries (1p, 2p, 3p), conjunctive queries (2i, 3i, ip, pi) and disjunctive queries (2u, up). For a given query $q$, the three sets of answers are obtained by performing subgraph matching over three graphs: $\llbracket q \rrbracket_{train}$ from $\mathcal{G}_{train}$, $\llbracket q \rrbracket_{valid}$ from $\mathcal{G}_{valid}\backslash \mathcal{G}_{train}$ and $\llbracket q \rrbracket_{test}$ from $\mathcal{G}_{test}\backslash \mathcal{G}_{valid}$. Therefore, it does not spoil the train/dev/test split in the original dataset.  We use $\llbracket q \rrbracket_{train}$ for fine-tuning training, validate over $\llbracket q \rrbracket_{valid}$ and test over $\llbracket q \rrbracket_{test}$. Such design allows us to test if the model can predict non-trivial answers unable to obtain by traversing the given KGs. 

\vpara{Evaluation Protocol.}We follow the evaluation protocol in Query2Box~\cite{ren2019query2box} including the filtering setting~\cite{NIPS2013_1cecc7a7} and the same metrics calculation. In filtering setting, since the answers to most queries are not unique, when calculating the rank of one answer $a$, we rule out other correct answers. And we use Hits at K(Hits@Km) as metrics following other baselines (which is different from Hits@K, as it requires to first average on Hits@K per query and then average over all queries, see more details in Appendix~\ref{app:eval}).

\subsection{Main Results}
We compare our model with various approaches, including GQE\cite{hamilton2018embedding}, Q2B\cite{ren2019query2box}, BiQE\cite{kotnis2021answering}, EmQL~\cite{sun2021faithful}, and state-of-the-art baseline CQD \cite{arakelyan2021complex}, covering methods based on embeddings and sequence encoders. Detail descriptions can be found in the Appendix~\ref{app:baseline}.

We present the results of Hits@3m for all query types over FB15k-237 and NELL995 in Table~\ref{tab:main}. On average, \method obtains the best results over both datasets. In contrast with BiQE, which is also transformer-based, our model achieves $6.4\%$ (18.6\% relative) and $5.7\%$ ($19.5\%$ relative) improvement on NELL995 and  FB15k-237  respectively on average without union operation. It demonstrates our model's superiority in architecture for higher capacity and better generalizability.
Besides, our \method can conduct union operations, supporting the complete set of EPFO queries.  

Compared to the previous state-of-the-art CQD, we obtain $2.4\%$ ($6.4\%$ relative) and $3.5\%$ ($12.1\%$ relative) improvement over NELL995 and FB15k-237 respectively. 
\method consistently performs better or comparably over most types of logic queries.
Especially for out-of-domain reasoning queries on FB15k-237, \method outperforms CQD by a significant margin, demonstrating pre-training's superior capability in helping model gaining more out-of-domain generalizability.

\method's major deficiency lies in 1p query identical to conventional knowledge graph completion problem. On one hand, this is probably because in our training objectives, we focus more on complex queries with multiple entities and relations instead of simple link prediction; on the other hand, \method's multi-layer architecture may be unfriendly to queries with fewer contexts. We believe there is plenty space to improve \method's architecture for better performance on 1p queries, and we will leave it for future research.

\begin{figure*}[t]
    \centering
    \includegraphics[width=\textwidth]{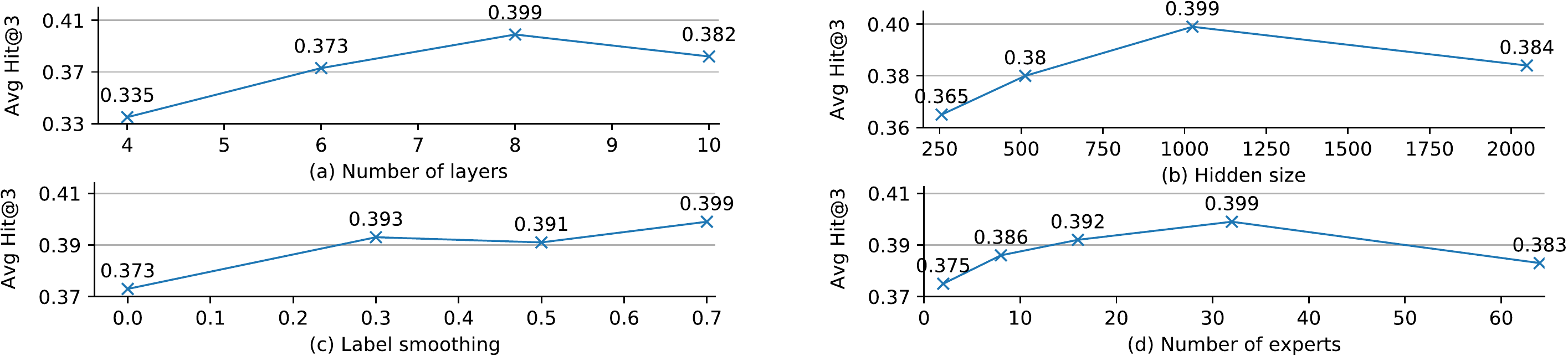}
    \caption{Ablation experiments \textmd{on NELL995 dataset (Hits@3m). The default setting is: 8 layers, 1024 for hidden size, $\alpha=0.7$ for label smoothing, and 32 for number of experts.} }
    \label{fig:ablation}
    \vspace{-2mm}
\end{figure*}

\subsection{Ablation Study}
\vpara{Adopted Strategies.}
We analyze the necessity of two-stage pre-training and the function of fine-tuning through ablation study on both datasets. We compare some \method's variants to demonstrate the contributions of the two stages in pre-training and the fine-tuning. The result is shown in Table~\ref{tab:ablation}. 

\begin{table}[t]
    \centering
    \renewcommand\tabcolsep{3pt}
    \caption{Ablation on certain pre-training \& fine-tuning strategies adopted (Hits@3m).}
    \begin{tabular}{lcc}
        \toprule[1.2pt]
        & FB15k-237 & NELL995\\ \midrule
        \method (Stage 1 + Stage 2) & \textbf{0.336} & 0.395 \\ \midrule
        \quad -only Stage 1 in pre-training & 0.308 & 0.307\\
        \quad -only Stage 2 in pre-training & 0.307 & \textbf{0.399}\\ \midrule
        \quad -w/o fine-tuning & 0.301 & 0.368\\
        \quad -w/o pre-training & 0.262 & 0.288\\
        \bottomrule[1.2pt]
    \end{tabular}
    \label{tab:ablation}
\end{table}



First, by comparing the result of \method, without fine-tuning and without pre-training, we observe that \method without pre-training decreases sharply by  $7\%-11\%$, while \method without fine-tuning decreases by around $3\%$.
It indicates that both of them are necessary to the success of \method, while pre-training has a decisive impact.
Even without the downstream training, \method with only pre-training can achieve state-of-the-art on FB15k-237 (higher than CQD~\cite{arakelyan2021complex}'s $29.0\%$).

Furthermore, different pre-training stage may have different inductive bias towards the underlying logic. \method-only Stage 1 in pre-training assumes dense correlation between pairs of entities, while \method-only Stage 2 in pre-training assumes that the relations among entities follow the chain-rule. \method takes both logics into consideration. We observe that \method obtains the best result, while \method-only Stage 2 in pre-training has equally good performance on NELL995 dataset. It reflects some nature of the two datasets: NELL995 is sparser than FB15k-237 with weaker relations within clusters and stronger connections in chains.

\vpara{MoE Efficiency.}
We also present a speed comparison across different number of experts among pre-training, fine-tuning and inference stages (Cf. Table~\ref{tab:moe_speed}) with a batch size of 64 in both training and inference.
We observe that MoE can significantly cutting down computation cost while increasing the model capacity.
In the case of 32 experts, MoE helps to enlarge \method for 1600\% but only adds to 11.6\%-38.7\% computational cost.
Theoretically speaking, the major efficiency loss should be blamed on the smaller computing granularity: GPU kernels usually prefer and compute faster for square-shaped inputs. 
More experts in MoE in fact reduce the input granularity per expert, and thus results in the observed speed drop.

\begin{table}[t]
    \centering
    \caption{Training and inference time per step (ms) along with different number of experts using per step batch size 64.}
    \renewcommand\arraystretch{1.1}
    \renewcommand\tabcolsep{7pt}
    \begin{tabular}{@{}lcccc@{}}
    \toprule[1.2pt]
     \multirow{2}{*}{} & \multicolumn{4}{c}{Number of experts}\\\cmidrule(l){2-5}
     \multicolumn{1}{l}{} & 2 & 8 & 16 & 32  \\ 
     \midrule
     Pre-training stage 1 &41.46 & 42.02 & 45.38 & 49.30 \footnotesize(+18.9\%) \\
     Pre-training stage 2 &17.37 & 18.83 & 19.78 & 24.09 \footnotesize(+38.7\%) \\
     Fine-tuning & 32.41 & 32.91 & 36.46 & 37.47 \footnotesize(+15.6\%)  \\
     Inference & 12.86 &  13.36 & 13.85 & 14.35 \footnotesize(+11.6\%)  \\
     \bottomrule[1.2pt]
\end{tabular}
    \label{tab:moe_speed}
    \vspace{-3mm}
\end{table}

\begin{table*}[t]
	\centering
    \caption{Case study on 2p examples.
    \textmd{\textbf{(a) Interpretability.}\label{tab:interpret} Filling a certain entity as the tail prediction, we use \method to predict the unknown intermediates to test its interpretability. \textbf{(b) All-direction Reasoning.} In 2p, intermediate and tail entities are masked.
    \method considers both relations when predicting intermediate node (marked \textbf{\textcolor{orange}{orange}}). If the second relation is masked (i.e., conventional autoregressive reasoning), intermediate's predictions are valid for the first relation, but no longer for the second (marked \textcolor{red}{red}). }}
    \renewcommand\tabcolsep{10pt}
    \renewcommand\arraystretch{1.10}
    
\begin{tabular}{cccccc}
\toprule[1.2pt]
\textbf{Case} & \textbf{Head}&&\textbf{Intermediate entity}&&\textbf{Tail}\\
&(Groundtruth)&&(Top 3 prediction)&&(Prediction)\\
\hline
Interpretability&{Elmira(U.S. City)}&{\textit{location in}}&{U.S.}, {Canada}, {UK}&\textit{in country}&Davidson College\\ 
\midrule
\multirow{2}{*}{\makecell[c]{All-direction\\Reasoning}}&\multirow{2}{*}{\textit{Salt}(2010 film)}&\multirow{2}{*}{\textit{has subject}}&\textbf{\textcolor{orange}{CIA}}, Espionage, PTSD&\textit{employs}&George Bush\\ \cmidrule{4-4} \cmidrule{6-6}
&&&\textcolor{red}{Serial killer}, Espionage, PTSD&\textit{(masked)}&-\\
\bottomrule[1.2pt]
\end{tabular}

    \label{tab:2p_case}
    \vspace{-2mm}
\end{table*}

\vpara{Hyperparameters.}
In this section, we conduct further ablation experiments on hyper-parameters including the number of layers, hidden size, the value of label smoothing and the number of experts 

\begin{itemize}[leftmargin=*,itemsep=0pt,parsep=0.2em,topsep=0.3em,partopsep=0.3em]
\item \textbf{Number of layers:} 
In Figure~\ref{fig:ablation} (a), we test the performance of different number of layers.
It turns out that a sufficient and proper depth is essential for \method (8 in our case), which can be attributed to the relation between depth and the size of receptive field.
A too deep GNN may suffer from over-smoothing and training instability and causes a performance drop.

\item \textbf{Hidden size:} 
Figure~\ref{fig:ablation}(b) studies the influence of hidden size. 
We observe that enlarging hidden size helps to capture massive EPFO reasoning patterns, but a too-large one can harm the performance, as it may suffer from the similar training difficulty witnessed in KGE-based reasoners~\cite{arakelyan2021complex}.

\item \textbf{Label smoothing:} 
Label smoothing is crucial to \method's pre-training, as it mitigates the bias in pre-training data due to the on-the-fly sampling. 
Through experiments in Figure~\ref{fig:ablation}(c), the optimal value $\alpha=0.7$ for NELL dataset is exceptionally high in comparison with that in natural language processing ($\alpha=0.1$). 
It indicates that by training for long epochs (around 1000 epochs) with a rather large embedding size (1024), our model is prone to overfitting in pre-training.  
We also observe that exclusion of label smoothing will yield a considerable performance decrease by $2.5\%$. 

\item \textbf{Number of experts:} 
We evaluate the performance of our model with different number of experts in Figure~\ref{fig:ablation}(d).  
The correspondence of sparse nature in KG reasoning and mixture-of-experts is one key point in our model. 
Compared with vanilla transformer (number of experts=2), our model achieves a boost of $2.3\%$ in the optimal case (number of experts=32), indicating the necessity of high-capacity models in EPFO challenge;
while too many experts (number of experts=64) may be counterproductive.

\end{itemize}

\subsection{Case Study: Interpretability and All-direction Reasoning}

\vpara{Interpretability.}In masked querying, because the model does not know the queried nodes in advance, it enforces the model to produce plausible predictions for \emph{every} node in the graph. Such property implies the interpretability of the results. The embeddings of intermediate variables can be fed to the prediction layer to give a result. This provides insights to how the model produced the answer, and benefits manual analyses as well. 

In Table~\ref{tab:interpret} (Interpretability), for example, in query ``Which school is in the same country as Elmira?'' (formally, $?A, \exists E, \text{location in}(Elmira, E)\land \text{in country}(E, A)$), the predicted answer is ``Davidson College''. To verify this answer, we would like to know what the result of $E$ is. Put the out embedding of $E$ into decoder, it will give ``U.S.'' as the top prediction, which is indeed a valid intermediate entity. More cases concerning interpretability for 3p queries are provided in Table~\ref{tab:3p_case}.

\vpara{All-direction Reasoning.}
Unlike almost all reasoner's step-by-step searching, \method is a transformer-based GNN that captures information from all directions simultaneously. As the second line in Table \ref{tab:2p_case} (All-direction Reasoning) shows, when the intermediate entity is required to fit the \textit{employs} relation, ``CIA'' is more suitable than ``Espionage'' or ``PTSD'' as it is an agency. Without the ability to read from tail to head, the model would suggest ``Serial killer'' instead, which is inadequate for future deductions.


}
\section{Related Work}
\label{sec:related}

Existing works on complex logical queries are based on KGEs~\cite{hamilton2018embedding,vilnis2018probabilistic,ren2019query2box,cohen2019neural,sun2021faithful,ganea2018hyperbolic,arakelyan2021complex}, sequence encoders~\cite{kotnis2021answering}, or rules~\cite{qu2020rnnlogic}. 
Most of them are based on traditional KGEs to adapt for complex logical reasoning with certain geological functions to manipulate the embedding spaces~\cite{hamilton2018embedding,ren2019query2box,sun2021faithful,ganea2018hyperbolic,ren2020beta} or logical norms over link predictors~\cite{arakelyan2021complex}. 
Graph Query Embedding (GQE) \cite{hamilton2018embedding} is proposed to use deep sets as a way for query intersection logic. 
Logical operators are reformulated as trainable geometric functions in the entity embedding space.
Q2B~\cite{ren2019query2box} embeds queries as boxes (i.e., hyper-rectangles), and the points inside which are considered as a potential answer entities of the query. 
Box Lattice \cite{vilnis2018probabilistic} is proposed to learn representations in the space of all boxes (axis-aligned hyper-rectangles). 
EmQL~\cite{sun2021faithful} proposes an embedding method with count-min sketch that memories the relations in training set well. 
CQD \cite{arakelyan2021complex} uses the previous neural link predictor ComplEx \cite{trouillon2016complex} as the one-hop reasoner and T-norms \cite{klement2013triangular} as logic operators. 
By assembling neural link predictors with T-norms, it translates each query into an end-to-end differentiable objectives. 
BetaE \cite{ren2020beta} utilizes Beta distribution for the embeddings. 
It takes the advantage of well-defined logic operation over distributions and turns the original logic operation over the embedding space into the one over the distribution space. 
These approaches can easily incorporate logic operations, but are hard to generalize well to unseen and more complicated queries, as their architectures of pure embeddings limit their expressiveness.

There have been efforts to employing advanced architectures for complex logical queries. BiQE~\cite{kotnis2021answering} introduces the Transformer and mask training, decomposing EPFO queries into sequences to fit the vanilla Transformer's input constraint. 
However, its nature to process sequences only allows it to answer queries in the shape of directed acyclic graphs (DAGs), which only cover a small set of all possible EPFO queries. 
In addition, it still follows the supervised training fashion and does not make the full use of the KGs.

Pre-training graph neural networks for better transferability and generalizability has recently aroused wide interest in graph community, inspired by the progress in language~\cite{devlin2019bert} and vision. 
Generally, these pre-training strategies can be categorized into two different self-supervised objectives~\cite{liu2021self}, that is,  generative pre-training~\cite{hu2020strategies,hu2020gpt} and contrastive pre-training~\cite{qiu2020gcc,you2020graph}. Nevertheless, they generally focus on pre-training graph neural networks on academic networks~\cite{zhang2019oag}, biochemical substances~\cite{sterling2015zinc}, or e-commerce product graphs~\cite{hu2020open}. 
Few efforts have been paid to the challenges of pre-training graph neural networks on knowledge graphs.
\section{Conclusion}
\label{sec:conclusion}

We present the Knowledge Graph Transformer (\method), a Transformer-based GNN, to pre-train for complex logical reasoning by leveraging the masked pre-training and fine-tuning approach. 
To adapt to graph structural data, we introduce the Triple Transformation and Mixture-of-Experts strategies for a high-capacity Transformer architecture, and propose a two-stage pre-training framework to gradually endow \method with the ability to transfer and generalize.
Extensive experiments and ablation study demonstrate the effectiveness of \method's architecture and pre-training methods.

Notwithstanding  the promising results, there exist limitations of \method that require future studies. 
First, \method is not competitive on 1p and 2u reasoning queries which are equivalent to the traditional KG completion problem), as we have not injected inductive biases known to be critical for GNNs on KGs (e.g., CompGCN~\cite{vashishth2019composition}). 
Second, the KG pre-training is limited to a single KG and does not benefit from cross-KG knowledge transfer.
It would be an interesting direction to fuse knowledge from multiple KGs via pre-training to gain further improvements on reasoning.
Third, as the recent work~\cite{saxena2022sequence} indicates, a sequence-to-sequence Transformer that leverages text labels of entities as inputs and outputs can serve as a promising architecture for KG completion in certain scenarios.
Finally, to jointly model the KG structure and text information in KG reasoning also remains an unsolved challenge. 

\section*{ACKNOWLEDGEMENT}
We thank the reviewers for their valuable feedback to improve this work. 
This work is supported by Technology and Innovation Major Project of the Ministry of Science and Technology of China under Grant 2020AAA0108400 and 2020AAA0108402, Natural Science
Foundation of China (Key Program, No. 61836013), and National Science
Foundation for Distinguished Young Scholars (No. 61825602).

\clearpage

\renewcommand\refname{REFERENCES}
\bibliographystyle{ACM-Reference-Format}
\balance
\bibliography{ref}

\clearpage
\appendix



\section{Sampling Method}

In the stage 1 of pre-training, we used random-walk based graph sampling strategies to sample dense and large subgraphs on the training set. We adopt LADIES \cite{zou2019layerdependent} sampling method and a variant of random walk called meta-tree sampling.

\vpara{Meta-tree Sampling.}
As a variant of random walk with restart, meta-tree sampling strategy does not necessarily restart from the target node every time. Instead, it restarts from any sampled node with equal probability. The change is inspired by the intention to sample more neighbors for each node rather than for the target node in a subgraph. In comparison with tuning the parameters of RW, meta-tree sampling provides a direct way to spread the \emph{density} across all the sampled nodes. Such a \emph{density spread} is essential for preventing overfitting to highly dense area. Besides, to sample a dense graph, we specify the restart probability to be $1.0$, which means it jumps to any sampled node after one step and continue. In this way,  meta-tress sampling balances between the width and the depth and prevents overfitting problem in our setting. Note that we do not sample a node more than once, so the sampled subgraph contains no cycle and forms a tree.

\section{Evaluation Protocol}
\label{app:eval}
To be more precise, we denote the whole entity set as $\mathcal{V}$, and the filtered entity set w.r.t qeury $q$ as $\mathcal{S}=\mathcal{V}\backslash \llbracket q \rrbracket_{\text{test}}$.
\begin{equation}
    \text{rank}_\mathcal{V}(a)= 1 +\sum_{x\in \mathcal{S}} \mathbb{1}[ \text{rank}(x)<\text{rank}(a)]
\end{equation}
The metric Mean Reciprocal Rank (MRR) is caculated as $\frac{1}{\text{rank}}$ and Hits at K(Hits@Km) is calculated as $\mathbb{1}[rank<K]$. In this paper, we report the Hits@3m metric, the frequency of the correct answer to appear in top 3 ranking. The metric of a query is an average of the metrics over all its predicted answers, and the metric of a type of query is an average of the metrics over all queries of this type.

\section{Reproducibility}
\label{app:Reproducibility}
In this section, we describe the experiment setting in more details. 
\vpara{Label Smoothing.}
Label smoothing is a technique of regularization. It utilizes soft one-hop labels instead of hard ones to add noise in the training, reduces the weights of true labels when calculating the training loss and thus prevents overfitting when training. Suppose $y_h$ is the original true label, $y_{ls}$ is the smoothed label, $\alpha$ is the parameter of label smoothing, $K$ is the number of classes. 
\begin{equation}
    y_{ls} = (1-\alpha) y_{h} + \frac{\alpha}{K}
\end{equation}
Since the data used in pre-training is the full graph, much larger than the query data in fine-tuning, we add label smoothing only in pre-training. We used $\alpha=0.1$ for FB15k-237 and $\alpha=0.7$ for NELL. Note that because the data used in fine-tuning is limited, we do not add label smoothing in fine-tuning. 

\vpara{Sampling Parameters.}
We leverage several sampling methods in the two pre-training stages where sampling ratio makes a difference to the final performance. In pre-training stage 1, for FB15k-237, the ratio between meta-tree sampling and LADIES sampling is $1:1$; for NELL, we only use meta-tree sampling. In pre-training stage 2, the ratio between chain-like meta-graphs and branch-like meta-graphs can be various. We applied grid search within the range $[1:20, \ldots 1:2, 1:1, 2:1, \ldots 20:1]$ and set the ratio to be $4:1$ for FB15k-237 and $10:1$ for NELL. Besides, in the stage 1 of pre-training, for each sampled subgraphs, we keep $80\%$ of the induced edges between nodes and limit the number of nodes in each subgraph within $[8, 16]$. In the stage 2 of pre-training, we do not add induced edges and limit the size of meta-graphs to be less than 4. \\
In stage 2 pre-training, we sample two patterns of small graphs for refinement, chain-like subgraphs and branch-like subgraphs. We first select a target node from all the nodes uniformly at random. For the chain-like subgraph sampling, we sample a chain by the Markov process. In each step, we sample a neighbor of the current node independently. It is allowed to sample a node more than once in a chain. For the branch-like subgraph sampling, we sample multiple neighbors of the target node. We drop those sampled graphs whose target node has no or only one neighbor and prevent sampling the same neighbors multiple times to avoid meaningless cases. 

\vpara{Mixture-of-Experts (MoE).}
We have already discussed the importance of MoE in the setting of multiple patterns reasoning especially when the FFN is sparsely activated. Note that we use Pre-LN (Pre-Layer Normalization, which refers to placing the layer normalization in the residual connections and appending an auxiliary layer normalization before the final linear decoder~\cite{xiong2020layer}) instead of Post-LN (Post-Layer Normalization, which refers to vanilla Transformer's design of placing the layer normalization between the residual blocks) in MoE, as evidences show that Pre-LN converges faster and more robustly compared to Post-LN~\cite{xiong2020layer}. For the number of experts, using 2 experts of MoE means the original setting of FFN. We applied grid search over the number of experts within $[2, 4, 8\ldots32, 64]$ and select 32 as the number of experts. We implement the MoE strategy using the FastMoE\footnote{\url{https://github.com/laekov/fastmoe}}~\cite{he2021fastmoe}, an open-source pytorch-based package for MoE with transformers.

\vpara{Training Parameters.}
To enhance the generalization ability of our model, we add some noise to the \emph{mask} in pre-training stage 1, following the strategies in BERT. For the nodes to be masked,
\begin{itemize}[leftmargin=*,itemsep=0pt,parsep=0.2em,topsep=0.3em,partopsep=0.3em]
    \item $80\%$ of the time, we replace the node with [mask] token;
    \item $10\%$ of the time,  we keep the node unchanged;
    \item $10\%$ of the time, we replace the node with a random node. 
\end{itemize}
Such masking ratio forces our model to keep a contextual representation for every node and thus learn the neighborhood information.\\
We train our model with batch size 258 in pre-training and 12288 in fine-tuning. We use AdamW\cite{loshchilov2019decoupled} with $lr=1e-4, \beta_1=0.9, \beta_2=0.999$, exponential decay rate of 0.997. We also use a dropout probability of 0.1 every layer.

\begin{table*}[t]
    \centering
    \caption{Case study on 3p examples (Interpretability).
    \textmd{Filling a certain entity as the prediction for tail, we use \method to predict the unknown intermediate node embedding to test its interpretability. Compared to 2p, 3p queries have one more intermediate entity.
    Each line correspond to a back-query and 
    the intermediate entities reacts to the change of tail entities.
    }}
    \renewcommand\tabcolsep{7.5pt}
    \renewcommand\arraystretch{1.1}
    \begin{tabular}{
p{2.2cm}<{\centering}
p{1.0cm}<{\centering}
p{2.2cm}<{\centering}
p{2.0cm}<{\centering}
p{2.6cm}<{\centering}
p{1.7cm}<{\centering}
p{1.8cm}<{\centering}
}
\toprule[1.2pt]
\textbf{Head}&&\textbf{Intermediate 1}&&\textbf{Intermediate 2}&&\textbf{Tail}\\
(Ground truth)&&(Back-queried)&&(Back-queried)&&(Prediction)\\
\hline
\multirow{3}{*}{TV5 (TV network)}&\multirow{3}{*}{\textit{leader title}}&President&\multirow{3}{*}{\textit{company with title}}&Verizon&\multirow{3}{*}{\textit{region in}}&New York\\
&&President&&eBay&&California\\
&&President&&Ford Motor&&Michigan\\
\hline
\multirow{3}{*}{Priyanka Chopra}&\multirow{3}{*}{\textit{gender}}&Female&\multirow{3}{*}{\textit{risk factors}}&Hypothyroidism&\multirow{3}{*}{\textit{risk factors}}&Depression\\
&&Female&&Cirrhosis&&Peritonitis\\
&&Female&&Cirrhosis&&Liver failure\\
\hline
\multirow{3}{*}{Clarinet}&\multirow{3}{*}{\textit{role}}&Paquito D'Rivera&\multirow{3}{*}{\textit{Profession}}&Musician&\multirow{3}{*}{\textit{specialization of}}&Artist\\
&&Imogen Heap&&Singer-songwriter&&Musician\\
&&Harry Shearer&&Author&&Writer\\
\bottomrule[1.2pt]
\end{tabular}

    \label{tab:3p_case}
\end{table*}

\section{Combinatorial Fine-Tuning}
\begin{table}[h]
    \small
    \centering
    \caption{Preliminary experiments on combinatorial strategies in the single-task tuning after multi-task tuning on NELL995. \textmd{Base case refers to the results after multi-task tuning. Bold results indicate the most effective single-task tuning method for each query. In this version not all the strategies reported in the final version are applied.}}
    \begin{threeparttable}
    \renewcommand\tabcolsep{2.5pt}
    \renewcommand\arraystretch{1.1}
    \begin{tabular}{@{}lccccccccc@{}}
    \toprule[1.2pt]
     \multirow{2}{*}{} & \multicolumn{5}{c}{In-domain} & \multicolumn{4}{c}{Out-of-domain}\\\cmidrule(l){2-6} \cmidrule(l){7-10}
     \multicolumn{1}{l}{} & 1p & 2p & 3p & 2i & 3i & ip & pi & 2u & up  \\ 
     \midrule
     Base & 0.605 &	0.345 &	0.277 &	0.373 &	0.521 &	0.164 &	0.252 & 0.428  & 0.265 \\ \midrule
     1p & 0.609 &	0.351 &	0.282 &	0.377 & \textbf{0.523} &	0.178 &	\textbf{0.267} & \textbf{0.432}  & 0.258 \\
     2p & 0.613  &	0.355 &	0.278 &	0.379 &	0.515 &	0.166 &	0.258 & 0.432  & 0.269 \\
     3p & 0.609 &	0.328 &	0.285 &	0.378 &	0.522 &	0.162 &	0.253 & 0.425  & 0.262 \\
     2i & 0.594 &	0.339 &	0.273 &	\textbf{0.390} &	0.521 &	0.155 &	0.249 & 0.427  & 0.254 \\
     3i & 0.590 &	0.334 &	0.274 &	0.356 &	0.522 &	0.152 &	0.237 & 0.428  & 0.255 \\
     1p, 2p & 0.612 &	\textbf{0.358} &	0.281 &	0.368 &	0.499 &	\textbf{0.183} &	0.263 & 0.431  & \textbf{0.272}\\
     1p, 3p & \textbf{0.615} &	0.348 &	0.287 &	0.362 &	0.491 &	0.169 &	0.253 & 0.428 & 0.263 \\
     1p, 2p, 3p & 0.611 &	0.351 &	\textbf{0.288} &0.363 &	0.494 &	0.183 &	0.257 & 0.431 & 0.266 \\
     \bottomrule[1.2pt]
\end{tabular}
    \end{threeparttable}
    \label{tab:multi-task}
\end{table}

In the fine-tuning stage, we conduct multi-task fine-tuning followed by single-task fine-tuning. Multi-task fine-tuning refers to fine-tuning the pretrained model with all of the query-answer training sets, including $1p, 2p, 3p,2i,3i$. Single-task fine-tuning follows multi-task fine-tuning by tuning with only part of the query-answer training sets, either a single query's set or a combination of several query's sets. To compare different combinations in single-task fine-tuning, we just pick an arbitrary checkpoint in pretraining stage with preliminary result, do multi-task fine-tuning and then show different combinations of single-task fine-tune.

\section{Comparison Methods}
\label{app:baseline}
In this section, we briefly go through the compared baselines for reference.

\begin{itemize}[leftmargin=*,itemsep=0pt,parsep=0.2em,topsep=0.3em,partopsep=0.3em]
    \item \textbf{GQE~\cite{hamilton2018embedding}} utilizes deep set for intersection (conjunctive) logical operation as learned geometric functions in this space.
    \item \textbf{Q2B~\cite{ren2019query2box}} embeds queries as hyper-rectangles and answers as points inside the rectangles.
    \item \textbf{EmQL~\cite{sun2021faithful}} utilizes count-min sketch for query embedding in order to be more faithful to deductive reasoning. It focuses on deductive reasoning which does not require generalization.
    \item \textbf{BiQE~\cite{kotnis2021answering}} utilizes bi-directional attention mechanism and positional embedding to capture interactions within a graph. 
    \item \textbf{CQD~\cite{arakelyan2021complex}} uses ComplEx as one-hop reasoner and various T-norms as logic operators. It provides end-to-end differentiable objective by uniting one-hop reasoners and logic operators. 
\end{itemize}

\begin{table}[t]
    \centering
    \caption{Dataset statistics and details of splitting.}
    \small
    \begin{tabular}{ccccccc}
         \toprule[1.2pt]
         Dataset & \#Ent. & \#Rel. & \#Edges & Train & Valid & Test\\
         \midrule
         FB15k-237 & 14,505  & 237 & 310,079 & 272,115 & 17,526 & 20,438\\
         NELL995 & 63,361 & 200 & 142,804 & 114,213 & 14,324 & 14,267\\
         \bottomrule[1.2pt]
    \end{tabular}
    \label{tab:dataset}
\end{table}

\begin{table}[t]
    \centering
    \small
    \caption{Splitting statistics of query-answer dataset.}
    \renewcommand\tabcolsep{3.5pt}
    \renewcommand\arraystretch{1.1}
    \begin{tabular}{cccccccc}
         \toprule[1.2pt]
         \multirow{2}{*}{Dataset} & \multicolumn{2}{c}{Train}  & \multicolumn{2}{c}{Valid} & \multicolumn{2}{c}{Test}\\
         \cmidrule(l){2-3} \cmidrule(l){4-5} \cmidrule(l){6-7}
         & 1p & others & 1p & others & 1p & others\\ 
         \midrule
         FB15k-237 & 149689 & 149689 & 20101 & 5000 & 22812 & 5000\\
         NELL-995 & 107982 & 107982 & 16927 & 4000 & 16927 & 4000\\
         \bottomrule[1.2pt]
    \end{tabular}
    \label{tab:split}
\end{table}

\section{Statistics of Datasets}
We provide the information of the splitting of original datasets in Table \ref{tab:dataset} and the splitting of the query-answer datasets in Table \ref{tab:split}.

\end{document}